\pdfoutput=1

\documentclass[11pt]{article}

\usepackage[dvipsnames]{xcolor}
\usepackage[preprint]{acl}

\usepackage{times}
\usepackage{graphicx}
\usepackage{amssymb}
\usepackage{latexsym}
\usepackage{booktabs}
\usepackage{adjustbox}
\usepackage{multirow}
\usepackage{amsmath}
\usepackage{pifont} 
\usepackage{colortbl}

\definecolor{myblue}{RGB}{14, 121, 178}
\usepackage[T1]{fontenc}

\usepackage[utf8]{inputenc}

\usepackage{microtype}

\usepackage{inconsolata}

\usepackage{graphicx}
\usepackage{enumitem}
\title{{\scalebox{1}[1]{\includegraphics[width=1.0cm]{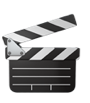}}}Clapper: Compact Learning and Video Representation in VLMs}

\author{
 \textbf{Lingyu Kong\textsuperscript{1}},
 \textbf{Hongzhi Zhang\textsuperscript{2}},
 \textbf{Jingyuan Zhang\textsuperscript{2}},
 \textbf{Jianzhao Huang\textsuperscript{2}},
\\
 \textbf{Kunze Li\textsuperscript{3}},
 \textbf{Qi Wang\textsuperscript{2}},
 \textbf{Fuzheng Zhang\textsuperscript{2}}
\\
\\
 \textsuperscript{1}University of Chinese Academy of Sciences,
 \textsuperscript{2}Kuaishou Technology,
 \textsuperscript{3}Xi’an Jiaotong University
\\
 \small{
   \textbf{Correspondence:} \href{mailto:konglingyu20@mails.ucas.ac.cn}{konglingyu20@mails.ucas.ac.cn}
 }
}

\begin{document}
\maketitle

\begin{abstract}
Current vision-language models (VLMs) have demonstrated remarkable capabilities across diverse video understanding applications. Designing VLMs for video inputs requires effectively modeling the temporal dimension (i.e. capturing dependencies across frames) and balancing the processing of short and long videos. Specifically, short videos demand preservation of fine-grained details, whereas long videos require strategic compression of visual information to handle extensive temporal contexts efficiently.
However, our empirical analysis reveals a critical limitation: most existing VLMs suffer severe performance degradation in long video understanding tasks when compressing visual tokens below a quarter of their original visual tokens.
To enable more effective modeling of both short and long video inputs, we propose \textbf{Clapper}, a method that utilizes a slow-fast strategy for video representation and introduces a novel module named TimePerceiver for efficient temporal-spatial encoding within existing VLM backbones.
By using our method, we achieves 13x compression of visual tokens per frame (averaging 61 tokens/frame) without compromising QA accuracy.
In our experiments, Clapper achieves 62.0\% on VideoMME, 69.8\% on MLVU, and 67.4\% on TempCompass, all with fewer than 6,000 visual tokens per video. The code will be publicly available on the homepage.
\end{abstract}

\section{Introduction}

Vision Language Models (VLMs) have achieved significant progress in understanding single images, high-resolution images \cite{wei2023vary,ye2023mplug,chen2024onechart}, and multiple images \cite{BLIP2, Flamingo, llavaonevision} over the past few years. However, representation and understanding of videos in VLMs remain relatively underexplored. As an extension of images, videos comprise sequences of frames that introduce an additional temporal dimension. Unlike multiple-image inputs, videos continuously extend over time, typically recorded at 30 frames per second (fps) or 24 fps, offering rich and dynamic information. This temporal richness also imposes substantially higher computational demands.
For instance, even when the video is sampled at a reduced rate of 1 fps, if we use 196 visual tokens per frame~\cite{llavaonevision, longva}, we would need to encode 110,760 tokens in VLMs for a 10-minute video.

Many VLM works have been proposed to model the temporal characteristic of video data for video understanding. Some methods~\cite{lin2023video,internvideo2} introduce dedicated video encoders specifically designed for video representation, while others~\cite{xu2024pllava,xu2024slowfast,vcgbench,llavaonevision} reuse the image encoders and apply spatial or temporal pooling to video frame features without adding extra training parameters.
%
Additionally, some research works aim to reduce the number of tokens required for video inputs.
For example, some methods~\cite{llamavid,chen2024videollm} compress video clips at extreme ratios, such as reducing 4 or 8 frames to 1–32 tokens. However, they may result in significant loss of detailed information, which undermines their performance in fine-grained video QA tasks such as VideoMME~\cite{videomme}. Aiming to achieve strong performance across a variety of Video QA benchmarks with different video lengths and question types, models must make trade-offs between video token compression rates and the preservation of various detailed information.
While high compression rates enable the model to accept longer inputs, which benefits long videos, the information loss introduced by such high compression often degrades QA performance on short videos.

Another critical issue we find is the lack of a fair evaluation setting of video understanding tasks. Many results in current video QA benchmarks fail to report the number of tokens used by the video, which can lead to unfair comparisons. This is because, for a given number of input frames, the more vision tokens used, the better the QA performance typically becomes. Even for the same model, increasing the number of input frames within certain limits can lead to significant performance improvements.

In this work, we aim to study
effective modeling and fair evaluation that can be applied to both short and long video
inputs.
For effective modeling, we propose a more effective and efficient approach by employing an image-based vision encoder combined with a TimePerceiver module to represent video content, allowing video compact learning within the VLM framework.
For fair evaluation, our goal is to provide the video QA benchmark results within a fixed vision token upper bound. These results not only advocate for fairer comparisons, but also offer more relevant insights for practical applications, where a balance between computational cost and accuracy must be achieved.

The contributions of this work are as follows:
\begin{itemize}[noitemsep]
    \item We propose \textbf{Clapper}, a competitive VLM for video understanding, which is capable of achieving strong performance using less than one third of the tokens required by previous state-of-the-art models.
    \item We introduce a novel video representation method that leverages a slow-fast strategy and the TimePerceiver module, enabling efficient and compact learning for VLMs.
    \item We explored the performance of current VLMs under a unified upper bound on video tokens, providing greater practical value for the application of video VLMs in real-world scenarios.
\end{itemize}

\section{Related Works}
\subsection{Video Representation}
For video inputs in VLM, frames are typically sampled at a fixed frame rate, such as 1 fps, or a fixed number of frames are force-sampled and fed into the model.
The VLMs are then revised to embed the video frames, producing an embedding that represents the video segment. For instance, InternVideo2~\cite{internvideo2} accepts a fixed input of 8 frames and outputs an embedding of size $C \times L = 3200 \times 2048$, $C$ is the channel dimension, and $L$ is the token length.

Some other models process frames individually through the image encoder, such as CLIP~\cite{radford2021clip} or SigLIP~\cite{zhai2023siglip}, to obtain their image embeddings. Ultimately, these visual inputs are then either passed through an MLP and directly concatenated with the text query, or processed using a Q-former structure, where an LLM performs video understanding tasks~\cite{BLIP2,damonlpsg2023videollama,llamavid}.
If the Q-former structure is used, we need to recalculate visual inputs for each query, which is inefficient for multi-turn dialogue. Direct concatenation of visual tokens is more simple, but it can result in excessively long sequences, which limit the performance and efficiency of the LLM.
Consequently, extensive research has focused on compressing visual tokens, and will be discussed in the following section.

\begin{figure*}[t]
  \centering
  \includegraphics[width=0.99\linewidth]{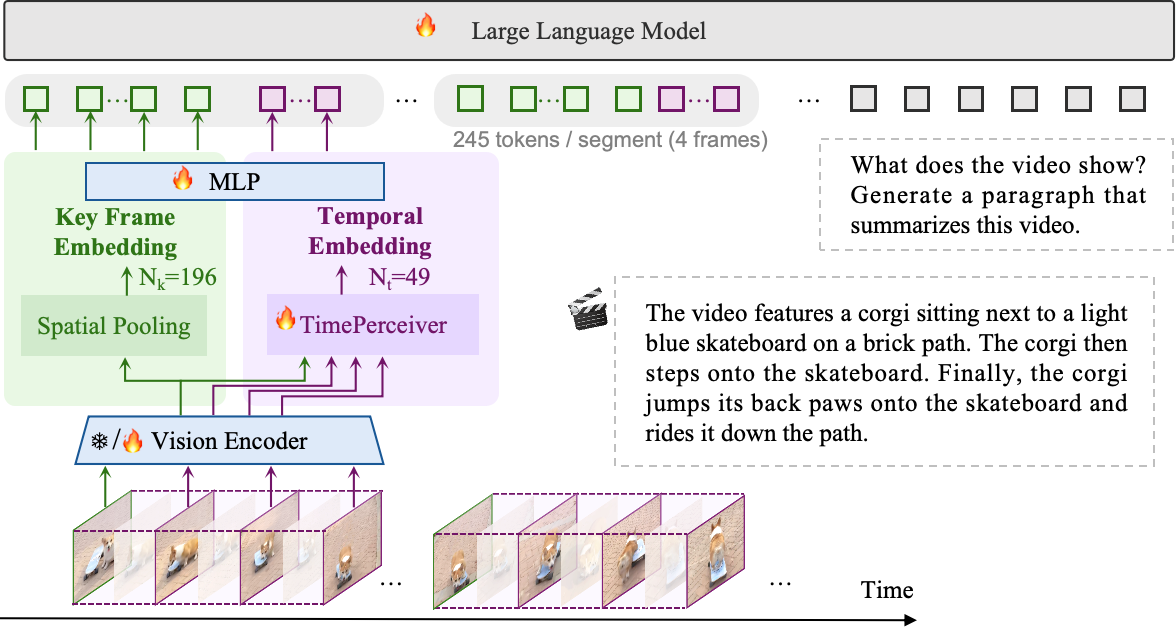} \hfill
  \caption {Architecture of Clapper. The model consists of a vision encoder, a TimePerceiver module, an MLP layer, and an LLM. Input videos are sampled and divided into segments, with each segment represented by a combination of high-resolution keyframe embedding and compressed temporal embedding.
  }
  \label{fig:Architecture}
\end{figure*}

\subsection{Token Compression}
Visual token compression have rapid advancements in VLMs with image inputs. For example, high-resolution images are essential for fine-grained perception tasks such as OCR, object grounding, and detailed information-based QA. To address the demands of high-resolution image inputs, dynamic high-resolution techniques have been proposed, which divide the high-resolution image into multiple small patches (referred to as "tails"). However, the increase in the number of tokens due to these multiple tails results in a rapid rise in memory usage and latency, which limits the practical applicability of VLMs. In response to the above issue, several models have introduced techniques such as pixel shuffle~\cite{internvl2,nvlm2024}, spatial pooling~\cite{llavaonevision,longva}, or bilinear interpolation~\cite{llavaonevision}, to reduce the number of tokens generated by these tails, typically achieving a reduction of approximately a quarter.

For video inputs, there are also some works that reduce the number of visual tokens of multiple frames from videos. PLLaVA~\cite{xu2024pllava} explored the effects of average pooling at varying strides in both spatial and temporal dimensions. Their experiments on MVBench~\cite{videochat2} and VCGBench~\cite{vcgbench} revealed that while 50\% spatial downsampling maintains performance levels, temporal pooling or more aggressive spatial compression leads to notable performance degradation. To mitigate resolution loss for critical frames, SlowFast~\cite{xu2024slowfast,llavavideo} adopts an asymmetric strategy where key frames undergo spatial downsampling by a factor of 2 (slow pathway) while other frames use downsampling by a factor of 4 (fast pathway). Meanwhile, alternative architectures have explored learnable compression mechanisms: MiniCPM~\cite{minicpm} and InternVideo2-HD~\cite{internvideo2} implement Perceiver-style cross-attention to project each frame's features into fixed-length tokens. However, experimental results indicate that these methods still suffer from non-negligible performance degradation and face challenges in further reducing token counts. LLaMA-VID~\cite{llamavid} achieves an exceptional compression by using context-aware techniques, where text queries and visual embeddings interact through cross-modal attention to produce two adaptive tokens per frame. However, this approach may not be optimal for all tasks, such as video captioning.

\section{Method}
\subsection{Architecture}
We introduce Clapper, a VLM for video understanding that achieves compressed video representation and aligns it with an LLM backbone. As shown in Figure~\ref{fig:Architecture}, unlike approaches that require extensive data resources to train a video foundation model from scratch or those that use image foundation models in a training-free manner, we adapt a pretrained image foundation model as the base.

Specifically, we employ SigLIP-448px/16~\cite{zhai2023siglip} as the vision encoder, which outputs a 784-dimensional embedding for each input image. For input videos, during training, we sample the video at 1 fps, resulting in a sequence of frames $I\in \mathbb{R}^{(T\times 448\times 448 \times 3)}$ with the length $T$. We sequentially divide the video into short segments, each consisting of 4 consecutive frames. The first frame of each segment is designated as the key frame, and its embedding is obtained by applying a spatial pooling with a stride of 2 to the original 784-dimensional embedding, resulting in a 196-dimensional representation. This key frame is crucial for preserving detailed spatial information, such as the attributes of the scene and the main subjects. It ensures that important visual details including the layout of the scene, the appearance of objects, and the positions of characters, are retained in the video representation.

We then introduce a trainable component, TimePerceiver, to effectively learn the temporal information within each segment. TimePerceiver takes 4 frames as input and outputs a 49-dimensional embedding representation. Ultimately, the video segment is represented by combining high-resolution keyframe information with highly compressed temporal information.
The keyframe captures the detailed spatial attributes, while TimePerceiver focuses on temporal dynamics, ensuring that both aspects are adequately represented in the final video embedding. Each video segment occupies a total of 245 tokens.

For the last video segment that may not have 4 frames, if it contains only one frame, it will have 196 tokens corresponding to the key frame. If it contains two or three frames, it still occupies 245 tokens. The visual tokens obtained from all processed video segments are concatenated in sequence and combined with the query to serve as the input to the LLM. In our experiments, we use Qwen2~\cite{qwen2} as our LLM backbone.

\begin{figure}[t]
  \includegraphics[width=\columnwidth]{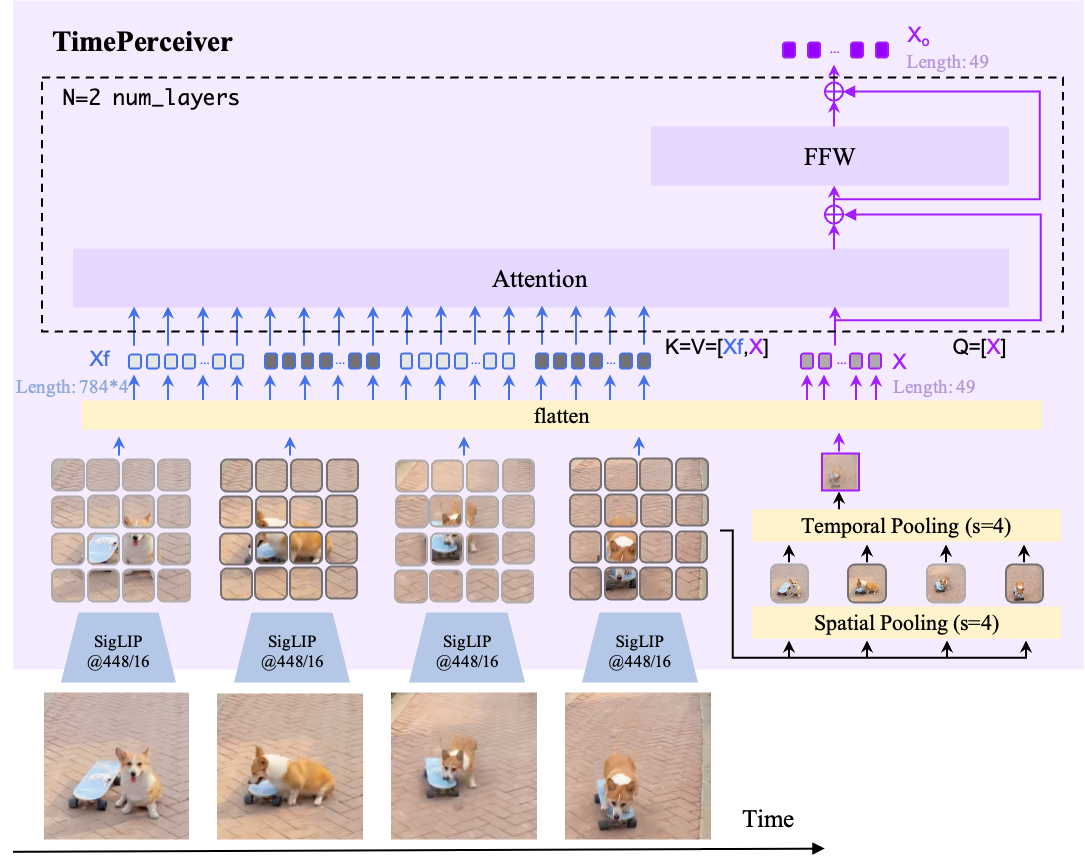}
  \caption{The \texttt{TimePerceiver} module processes 2-4 frames to generate a fixed number of temporal embedding outputs (49 in this paper).
  }
  \label{fig:TimePerceiver}
\end{figure}

The architecture of TimePerceiver is depicted in Figure~\ref{fig:TimePerceiver}. This module accepts a sequence of $T$ image features (in our approach, it typically receives 4 frames as input, though occasionally it may receive 2 to 3 frames) from the vision encoder, and it generates a fixed number of visual outputs (set to 49 in our method). For the input visual features $X_f \in \mathbb{R}^{(T \times L \times D)}$, we first apply spatial average pooling with a stride of 4, resulting in $X_s \in \mathbb{R}^{(T \times L/16 \times D)}$. Subsequently, we perform average pooling along the temporal dimension to obtain $X \in \mathbb{R}^{(1 \times L/16 \times D)}$. $L$ and $D$ represents the length and dimension of the visual features. In our method, $L$ is 784 and $D$ is 1152. By using $X$ as the input queries and cross-attending to the flattened visual features $X_f$, the model can focus on regions that change across the frames. Following the approach of Flamingo~\cite{Flamingo}, the keys and values are computed from the concatenation of $X$ and $X_f$. The number of output tokens from TimePerceiver is equal to the number of input queries. Our ablation studies demonstrate that employing such a TimePerceiver module yields superior performance compared to a standard Perceiver~\cite{Flamingo}.

\subsection{Training Recipe}
The training of Clapper is divided into two stages. In the first stage, TimePerceiver is trained to better model temporal motion information using video-caption pairs. During this stage, video-caption pair data is used for training. We first selected 178k video-caption pairs from LLaVA-Video~\cite{llavavideo}. These unedited videos, ranging from a few seconds to 3 minutes in length, offer detailed captions that describe various aspects of the content, capturing rich variations.
Additionally, we used OpenVid-1M~\cite{nan2024openvid}, a dataset for video generation. Videos in this dataset are trimmed to ensure scene consistency, and the captions are relatively concise.

In the second stage, we employ video instruction-tuning data to train the model's instruction-following capabilities. The training data includes LLaVA-Video-178K~\cite{llavavideo}, ActivityNet-QA~\cite{yu2019activitynet}, NExT-QA~\cite{nextqa}, PerceptionTest~\cite{perceptiontest}, and LLaVA-Hound-255K~\cite{zhang2024direct}, which together provide a total of 1.6 million video-language samples.

\section{Experiments}
We conducted evaluations across all benchmarks using LMMs-Eval~\cite{zhang2024lmms} to ensure standardization and reproducibility. To fairly compare with other leading VLMs that support video understanding, we primarily reused results from original papers. When results were not available, we integrated the models into LMMs-Eval and assessed them under consistent settings.

\subsection{Implementation Details}
We use  LLaVA-Onevision-SI~\cite{llavaonevision} with 7B parameters as our baseline model and finetune it for two stages. In the first stage, we freeze the weights of vision encoder and optimize the other parameters include the MLP layers, the TimePerveiver module and the language model. In the second stage, we finetune all parameters of the model on video instruction pairs. We train the model with a batch size of 128 for only 1 epoch in both two stages. The start learning rate is set to 2e-5. During training, videos are sampled at 1 fps, and for videos exceeding 96 seconds, we sample a fixed number of 96 frames at equal intervals.

\subsection{Overall Results}
We report the accuracy of Clapper on six popular and challenging multiple-choice video QA benchmarks. Compared to open-ended questions, multiple-choice questions are easier to evaluate and yield more stable results that are less influenced by the specific evaluation model. The benchmarks are introduced in the order of video duration, starting with TempCompass~\cite{tempcompass}, MVBench~\cite{videochat2} and PerceptionTest~\cite{perceptiontest}, which focus on short videos. LongVideoBench~\cite{longvideobench} and MLVU~\cite{mlvu} focus on long video understanding. Finally, VideoMME~\cite{videomme} covers a broad range of video understanding tasks. Most results are from original papers or benchmark leaderboards.

\paragraph{Overall comparison.}
\begin{table*}[!htb]
    \centering
\begin{adjustbox}{width=\linewidth,center}
\renewcommand{\arraystretch}{1.1}
\setlength{\tabcolsep}{1.5mm}
\begin{tabular}{lrrcccccccccccc}
\toprule  \multirow{2}{*}{\textbf{Model}} & \multicolumn{1}{c}{\multirow{2}{*}{\centering \textbf{Size}}} & \multicolumn{1}{c}{\textbf{\#Tokens./}} & \multirow{2}{*}{\textbf{\#Frames}} & \multirow{2}{*}{\textbf{\#Tokens}} & {\textbf{TempCompass}} & {\textbf{MVBench}} & {\textbf{PerceptionTest } }  &{\textbf{LongVideoBench}} & {\textbf{MLVU}} & \textbf{VideoMME}  \\ 
&&  \multicolumn{1}{c}{\textbf{Frame}} & & & M-Avg & Avg &Val&Val&M-Avg& Overall  \\
\rowcolor{gray!10} Avg. Duration & & & & & 13s & 16s& 23s  & 473s & 651s &  1010s  \\
\midrule
\textit{Proprietary Models} \\
GPT4-V~\citep{gpt4v} & - & -  & 1fps & - & - &  43.7 & - & 59.1  & 49.2 & 59.9/63.3 \\
GPT4-o~\citep{gpt4o} & - & -  & 1fps & - & 71.0 & 64.6 & - & 66.7  & 64.6 & 71.9/77.2  \\
Gemini-1.5-Pro~\citep{gemini} & - & - & 1fps & - & 67.1 & 60.5 & - & 64.0  & - & 75.0/81.3 \\
\midrule
\textit{Compression Ratio $\leq$ 4x} \\
IXComposer-2.5~\citep{xcomposer} & 7B & 400 & [32,64] & 26k & 61.3 & \underline{69.1} & 34.4 & -  & 58.8 & 55.8/58.8 \\
InternVL2~\citep{internvl2} & 8B & 256  & [8,16] & 4k & 65.6 & 65.8 & - & 54.6  & 64.0 & 54.0/56.9 \\
InternVL2.5~\citep{internvl2_5} & 8B & 256  & [16,32,48,64] & 16k & - & \textbf{72.0} & - & \textbf{60.0}  & 68.9 & \textbf{64.2}/66.9  \\
Kangaroo~\cite{kangaroo} & 8B & 256  & 64 & 16k & 62.5 & 61.1 & - & 54.8 & 61.0 & 56.0 / 57.6  \\
LongVILA~\citep{longvila} & 7B & 196  & 256 & 50k & - & 67.1 & 58.1 & 57.1  & - & 60.1/65.1 \\
LLaVA-Video~\citep{llavavideo} & 7B & 196  & 64 & 13k & 67.3 & 58.6 & \textbf{67.9} & \underline{58.2}  & \underline{70.8} & \underline{63.3}/\textbf{69.7}  \\
LLaVA-OneVision~\citep{llavaonevision} & 7B & 196  & 32 & 6k & 64.8 &56.7 & 57.1 & 56.3  & 64.7 & 58.2/61.5 \\
LLaVA-NeXT-Video~\citep{llavanextvideo} & 7B & 144 & 32 & 5k & 50.6 & 53.1 & 48.8  & 49.1 & - & \space\space\space-\space\space\space/46.5  \\
LongVA~\citep{longva} & 7B & 144 & [32,128] & 18k & 56.1 &  - & - & -  & 56.3 & 52.6/54.3 \\
LongLLaVA~\citep{longllava} & 9B & 144 & 128 & 18k & - & 49.1 & - & -  & -& 43.7/\space\space\space-\space\space\space \\
Qwen2-VL~\citep{qwen2vl} & 7B & Dyn & 1fps & - & \textbf{68.5} &67.0 & 62.3 & -  & - & \underline{63.3}/\underline{69.0}  \\
\midrule
\textit{4x $\textless$ Compression Ratio $\leq$ 16x} \\
MiniCPM-V 2.6~\cite{minicpm} & 8B & 96 & 64 & 6k & 66.3 & - & - & 54.9 & - & 60.9/63.7 \\
VideoLLaMA2~\citep{videollama2} & 7B & 72  & 16 & 1k & - & 54.6 & 51.4 & -  & 48.5 & 47.9/50.3 \\
VideoChat2-HD~\citep{videochat2} & 7B & 72  & 16 & 1k & 51.1 & 62.3 & - & -  & 47.9 & 45.3/55.7 \\
InternVideo2-HD~\citep{internvideo2} & 8B & 72 & 16 & 1k & - & 67.2 & 63.4 & -  & - & 49.4/\space\space\space-\space\space\space \\
LongVU~\citep{longvu} & 7B & 64 & 1fps & - & - & 66.9 & - & -  & 65.4 & \space\space\space-\space\space\space/60.6 \\
\rowcolor{blue!5} \textbf{Clapper~(Ours)} & 7B & 61 & 96 & 6k & \underline{67.4} & 60.3 & 66.5 &  55.6  & 69.8 & 62.0/64.6  \\
\rowcolor{blue!5} \textbf{Clapper~(Ours)} & 7B & 61 & 1fps & - & 67.3 & 59.6 & \underline{67.1} &  55.3  & \textbf{72.0} & 62.7/64.0  \\
\midrule
\textit{Extreme Compression Ratio} \\
LLaMA-VID~\citep{llamavid} & 7B & 2  & 1fps & - & 38.0 & 41.9 & 44.6 & - & 33.2 & 25.9/\space\space\space-\space\space\space  \\

\bottomrule
\end{tabular}
\end{adjustbox}
\caption{\textbf{Main results on multiple-choice video QA benchmarks.} M-Avg refers to the average score of multiple-choice QA tasks within each benchmark. 
The average tokens per frame and evaluation frames are shown for direct comparison.
In ``\#Tokens./frame'', Dyn denotes naive dynamic resolution. In ``\#Frames'', brackets indicate that scores are tested across multiple frame settings and reported as the highest value. The best and second-best results in open-source models are \textbf{bolded} and \underline{underlined}, respectively.}
\label{tab:overall_vqa}
\end{table*}
Table~\ref{tab:overall_vqa} shows the performance of Clapper compared to a wide range of VLMs that have leading performance on these benchmarks, including both proprietary models and open-source models within the 7B-9B parameter range. We explicitly display the number of frames used during evaluation and the average number of tokens per frame in the table. The open-source models are categorized based on their compression ratio into three groups: compression ratio $\leq$ 4x, 4x $\textless$ compression ratio $\leq$ 16x, and extreme compression ratio > 16x. The compression ratio (CR) is mathematically defined as:
\begin{equation}
CR = \frac{N_{\text{original}}}{N_{\text{compressed}}}
\end{equation}
where $N_{\text{original}}$ denotes the number of visual tokens generated by the vision encoder, and $N_{\text{compressed}}$ represents the number of compressed visual tokens actually used by the model. This metric quantifies the degree of token reduction achieved during the compression process, with higher values indicating more aggressive compression.

Our proposed Clapper achieves competitive results in the compression ratio > 4x category, obtaining the best performance on four benchmarks: TempCompass, PerceptionTest, MLVU and VideoMME. Specifically, Clapper achieves 67.4\% on TempCompass, 66.5\% on PerceptionTest, 69.8\% on MLVU and 62.0\% on VideoMME without subtitles. Compared to other models with similar performance, Clapper maintains a lower visual token overhead.

\begin{figure*}[h]
  \centering
  \includegraphics[width=1\linewidth]{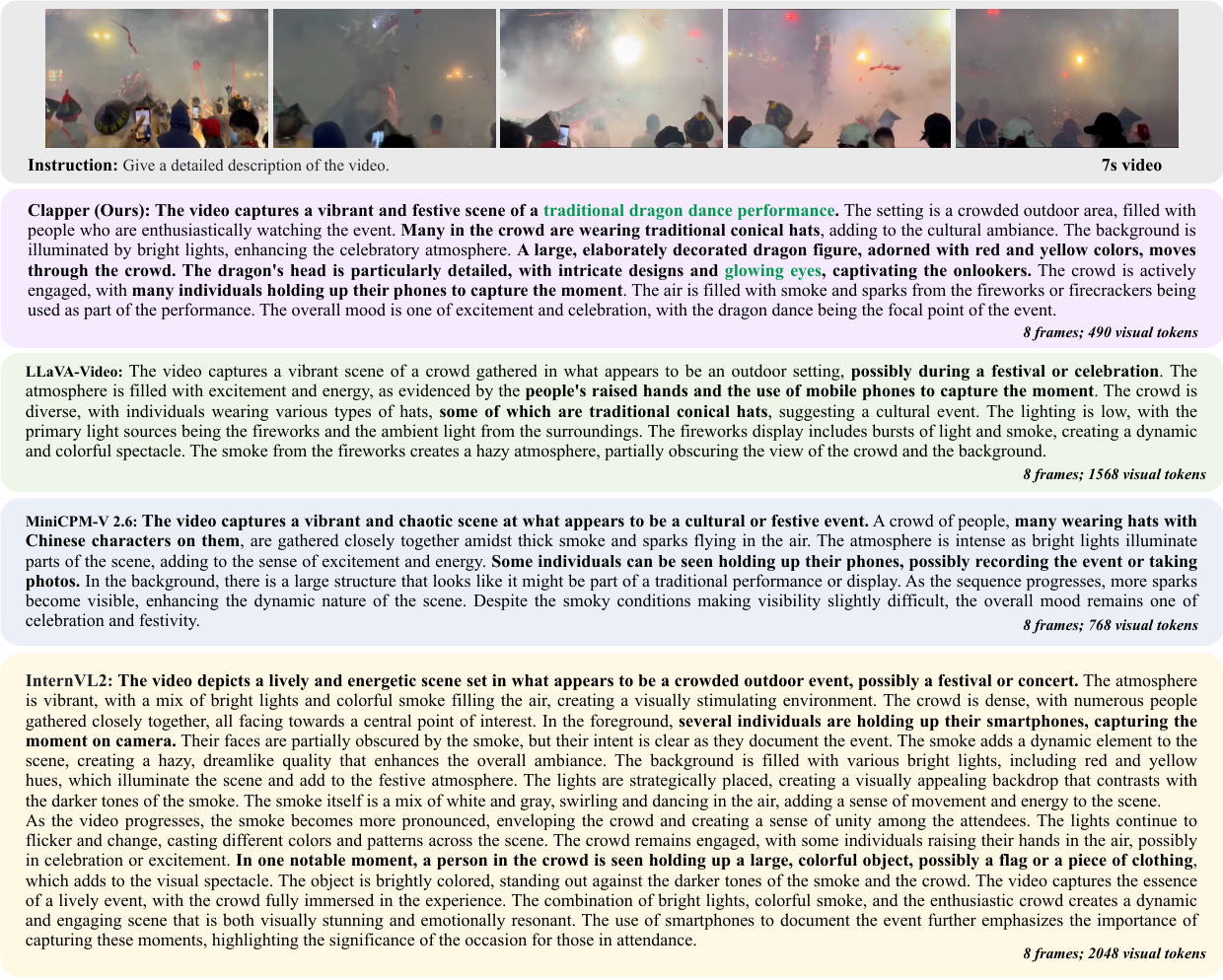}
  \caption {Comparison of video captioning results using Clapper and others. Key points are displayed in bold. Details uniquely captured by Clapper, which are absent in other models, are highlighted in bold green.}
  \label{fig:case_QA}
\end{figure*}

We also conducted a qualitative analysis of the model performance. As illustrated in Figure~\ref{fig:case_QA}, we compare the video captioning results of our proposed Clapper method with those of advanced models, including LLaVA-Video~\cite{llavavideo}, MiniCPM-V 2.6~\cite{minicpm}, and InternVL2~\cite{internvl2}. The figure also presents the visual tokens utilized by each method. Notably, Clapper is able to capture key points and details that are absent in the outputs of other models, as highlighted in bold green text. This demonstrates Clapper's superior ability to extract and summarize dynamic and detailed information, even with the minimal use of visual tokens.

\paragraph{Performance under different frames.}
\begin{figure}[t]
  \includegraphics[width=\columnwidth]{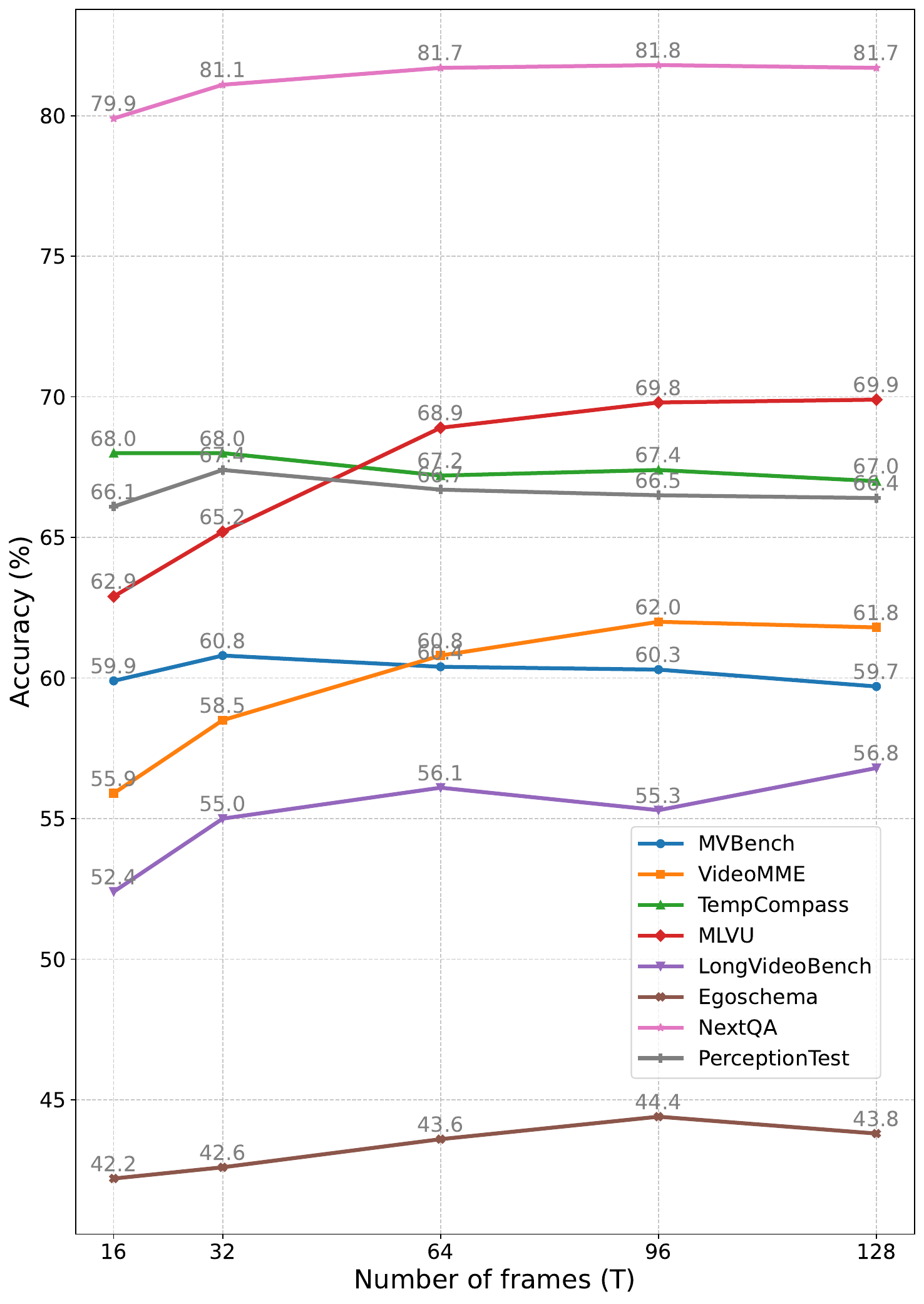}
  \caption{Performance of Clapper under different frames on the eight video QA benchmarks.
  }
  \label{fig:Result2}
\end{figure}
To explore how Clapper performs under varying visual token upper bounds, we adjusted the number of test frames for each video and observed the corresponding changes in performance across eight benchmarks. As illustrated in Figure~\ref{fig:Result2}, Clapper shows a steady improvement in MLVU performance as the number of input frames increases. On VideoMME, NextQA, LongVideoBench, and Egoschema, the overall trend also rises with increasing frames, though with some fluctuations. In contrast, performance in TempCompass, PerceptionTest, and MVBench shows little correlation with the number of input frames. These trends are highly correlated with the average video duration of each benchmark. When the average video length is less than the number of test frames, their performance does not benefit significantly from additional frames. In contrast, when the average video length is several times longer than the number of test frames, the models show a more substantial performance gain with increased frame sampling.

\paragraph{Performance under the same visual token upper bounds.}
To understand the trade-off between model accuracy and computational cost for practical deployment, we analyzed model performance by adjusting the number of input video frames to ensure all models operate under the same visual token upper bounds. Specifically, we set two token limits: 2k and 6k tokens. The number of input frames for each model was calculated by dividing these token limits by each model's tokens-per-frame number (\#Tokens/frame), as detailed in the \#Frames column in Table~\ref{tab:overall2}.

We selected VideoMME to compare different models under the same Visual Token Upper Bounds by varying the number of evaluation frames because it includes videos of varying lengths, categorized into short, medium, and long for evaluation. This facilitates our experimental observations.
The results are shown in Table~\ref{tab:overall2}. As can be seen, our proposed Clapper achieves the best performance within the 2k and 6k visual token limit, particularly excelling in short and medium-length videos. This highlights Clapper's ability to efficiently capture spatiotemporal information within limited token budgets. However, there is still significant room for improvement in understanding videos longer than 30 minutes within the 6k visual token limit. In the future, we aim to extend Clapper's strong performance to longer video durations.
\begin{table}[h]
\centering
\scalebox{0.74}{
\begin{tabular}{lc|ccc}
\toprule
\multirow{2}{*}{\textbf{Model}} & \multirow{2}{*}{\textbf{\#Frames}} & \multicolumn{3}{c}{VideoMME} \\
 & & Short & Medium & Long \\
\midrule
\textit{Visual Token Under 2k} \\
InternVideo2-HD & 32 & 53.7 & 43.0 & 40.1 \\
LLaVA-Video & 10  & 67.8 & 53.1 & 47.0 \\
\textbf{Clapper (Ours)} & 32  & 70.1 & 55.3 & 50.1   \\
\midrule
\textit{Visual Token Under 6k} \\
MiniCPM-V 2.6 & 64 & 71.3 & 59.4 & 51.8 \\
LLaVA-Video & 32  & 72.4 & 58.6 & 50.6 \\
\textbf{Clapper (Ours)} & 96  & 74.7 & 60.1 &  51.2  \\
\bottomrule
\end{tabular}
}
\caption{Detailed results on VideoMME (wo) with two fixed visual token upper bounds. 
Video lengths: Short (0–2 minutes), Medium (4–15 minutes), and Long (30–60 minutes).
}
\label{tab:overall2}
\end{table}

\subsection{Further Analysis}
To reduce experimental overhead, all models in the analysis section were trained using one-tenth of the full dataset used in Stage~2, which amounts to approximately 160k samples. This reduced dataset size allows for more efficient experiments while still providing sufficient data to evaluate the effectiveness of different components and strategies in our model design.

\paragraph{Analysis of video token compression.} To identify the optimal compression strategy that maintains the model performance while significantly reducing the token count, we conducted comprehensive experiments comparing different compression approaches for ablation. The Baseline method achieves a 4x compression of visual tokens by applying spatial pooling with a stride of 2 to all input frames. When using SigLIP@448px/16 as the vision encoder, the average number of tokens per frame in baseline is 196. To further reduce the token count from 4x to 16x compression, we compared four different compression strategies:
\begin{enumerate}
\item Temporal Pooling: Extend the baseline by incorporating temporal pooling with a stride of 4 along the temporal dimension.
\item Spatial Pooling: Build upon the baseline by applying an additional spatial pooling layer with a stride of 2.
\item Perceiver: Utilize a Perceiver architecture~\cite{Flamingo} with 49 randomly initialized learnable queries to aggregate information from every 4 frames.
\item TimePerceiver: Our method to integrate information from every 4 frames into 49 tokens.
\end{enumerate}

\begin{table}[t]
\centering

\scalebox{0.68}{
\begin{tabular}{ c c | c c c}
\toprule
Method & CR & MVBench & TempCompass & VideoMME \\
\midrule
Baseline & 4x & 55.4 & 63.1 & 59.1 \\
\midrule
Temporal Pooling & 16x & 56.3 \textcolor{OliveGreen}{(+0.9)} & 64.7 \textcolor{OliveGreen}{(+1.6)} & 56.9 \textcolor{BrickRed}{(-2.2)}\\
Spatial Pooling & 16x & 55.1 \textcolor{BrickRed}{(-0.3)} & 64.8 \textcolor{OliveGreen}{(+1.7)} & 58.4 \textcolor{BrickRed}{(-0.7)}\\
Perceiver & 13x & 56.5 \textcolor{OliveGreen}{(+1.1)} & 64.2 \textcolor{OliveGreen}{(+1.1)} & 56.5 \textcolor{BrickRed}{(-2.6)} \\
\textbf{TimePerceiver} & 13x & 57.2 \textcolor{OliveGreen}{(+1.8)} & 65.5 \textcolor{OliveGreen}{(+2.4)} & 59.3 \textcolor{OliveGreen}{(+0.2)} \\
\bottomrule
\end{tabular}
}
\caption{Comparison of video token compression strategies. CR stands for compression ratio. 
The performance differences show accuracy variations from the baseline.
}
\label{tab:ab1}
\end{table}

\noindent
In Perceiver and TimePerceiver, the first frame of every 4 frames is treated as a key frame and compressed using the baseline method. It is then concatenated with the generated 49 tokens to represent these 4 frames. Table~\ref{tab:ab1} shows the results. We evaluated these strategies on three video QA datasets with input frames uniformly sampled to 64. Temporal and Spatial Pooling achieve a 16x compression ratio, using 3k tokens per video, while Perceiver and TimePerceiver achieve a 13x compression ratio, using 4k tokens per video.

In benchmarks less sensitive to input frame count, like MVBench and TempCompass, methods that primarily compress temporal information, such as Temporal Pooling and Perceiver, perform better. Specifically, Temporal Pooling achieves a +0.9\% improvement on MVBench and a +1.6\% improvement on TempCompass. Perceiver improves by +1.1\% on both. However, on VideoMME, which is more frame-sensitive, both Temporal Pooling and Perceiver suffer significant performance drops, indicating substantial loss of temporal information. In contrast, Spatial Pooling achieves a +1.7\% improvement on TempCompass but shows a -0.3\% drop on MVBench and a -0.7\% drop on VideoMME. This suggests that MVBench and VideoMME may contain more fine-grained understanding questions, while TempCompass has lower requirements for spatial resolution.

Our proposed TimePerceiver method performs well across all three benchmarks, demonstrating effective compression in both spatial and temporal dimensions. Compared to the baseline with a 4x compression ratio, TimePerceiver achieves a 13x compression ratio while maintaining stable or even improved overall performance.

\paragraph{Impact of training designs.}
Experiments in this section were designed to evaluate how each training stage impacts the model performance, particularly the integration of the TimePerceiver module. We compared three setups: (1) a baseline model without TimePerceiver, (2) add TimePerceiver with direct training in the fine-tuning stage, and (3) add TimePerceiver with the full two-stage training strategy, stage 1 warm-up followed by stage 2 fine-tuning.

\begin{table}[h]
\centering

\scalebox{0.74}{
\begin{tabular}{c c | c c c}
\toprule
 Stage1 & Stage2 & MVBench & TempCompass & VideoMME \\
\midrule
\multicolumn{2}{c}{Baseline} & 55.4 & 63.1 & 59.1 \\
\midrule
\ding{55} & \checkmark & 55.9 \textcolor{OliveGreen}{(+0.5)} & 63.8 \textcolor{OliveGreen}{(+0.7)} & 58.3 \textcolor{BrickRed}{(-0.8)}\\
\checkmark & \checkmark & 57.2 \textcolor{OliveGreen}{(+1.8)} & 65.5 \textcolor{OliveGreen}{(+2.4)} & 59.3 \textcolor{OliveGreen}{(+0.2)} \\
\bottomrule
\end{tabular}
}
\caption{Ablation study on different training designs.
}
\label{tab:ab2}
\end{table}
Results in Table~\ref{tab:ab2} demonstrate that the inclusion of the TimePerceiver module, even with direct training, provides modest improvements over the baseline, with gains of +0.5 on MVBench and +0.7 on TempCompass. However, this setup shows a slight degradation of -0.8 on VideoMME, suggesting that direct training may not fully leverage the module's potential for video understanding. In contrast, the two-stage training strategy yields more significant improvements across all benchmarks, achieving +1.8 on MVBench, +2.4 on TempCompass, and +0.2 on VideoMME. These findings indicate that the two-stage training approach allows the TimePerceiver module to be more effectively trained for temporal information, leading to better overall performance on video understanding tasks. The warmup stage likely helps the model initialize and stabilize its learning of temporal features, while the fine-tuning stage refines these features for optimal performance. This highlights the importance of a carefully designed training strategy for integrating complex modules like TimePerceiver into video understanding models.

\section{Conclusion}
In this work, we proposed Clapper, an efficient video language model that demonstrates competitive performance across various benchmarks. Through the introduction of our TimePerceiver module, we successfully increased the compression ratio from 4x to 13x while preserving essential temporal and spatial information. This advancement allows Clapper to balance computational efficiency and accuracy, enabling faster inference and lower memory requirements. In the future, we aim to explore methods for further optimization of token representations to push compression boundaries even further. Additionally, we plan to extend our architecture's temporal modeling capacity to handle longer visual contexts.

\section{Limitations}
Clapper was trained at 1fps with a maximum of 96 frames, without additional training for length extrapolation. As a result, it is constrained by length extrapolation and the context length of large language models. For videos longer than 5 minutes, the model's performance may degrade as the video length increases. Additionally, Clapper has not undergone Reinforcement Learning from Human Feedback alignment training, which may lead to the generation of hallucinated outputs.

\bibliography{custom}




\end{document}